\documentclass{article} 
\usepackage{iclr2025_conference,times}

\usepackage{amsmath,amsfonts,bm}

\def\eqref#1{equation~\ref{#1}}

\def\1{\bm{1}}

\DeclareMathAlphabet{\mathsfit}{\encodingdefault}{\sfdefault}{m}{sl}
\SetMathAlphabet{\mathsfit}{bold}{\encodingdefault}{\sfdefault}{bx}{n}

\usepackage{hyperref}
\usepackage{url}

\usepackage{booktabs}
\usepackage{multirow}
\usepackage{graphicx}
\usepackage[normalem]{ulem}
\usepackage{comment}
\usepackage{algorithm}
\usepackage{algpseudocode}
\usepackage{adjustbox}
\usepackage{enumitem}
\useunder{\uline}{\ul}{}
\usepackage{caption}
\title{Realizing Video Summarization from the Path of Language-based Semantic Understanding}

\author{Kuan-Chen Mu\hspace{0.1em}, Zhi-Yi Chin\hspace{0.1em}, Wei-Chen Chiu \\
National Yang Ming Chiao Tung University\\
\texttt{\{kuanchenmu1999.cs10,joycenerd.cs09\}@nycu.edu.tw,walon@cs.nctu.edu.tw}
}

\iclrfinalcopy 
\begin{document}

\maketitle

\begin{abstract}
The recent development of Video-based Large Language Models (VideoLLMs), has significantly advanced video summarization by aligning video features—and, in some cases, audio features—with Large Language Models (LLMs). Each of these VideoLLMs possesses unique strengths and weaknesses. Many recent methods have required extensive fine-tuning to overcome the limitations of these models, which can be resource-intensive. In this work, we observe that the strengths of one VideoLLM can complement the weaknesses of another. Leveraging this insight, we propose a novel video summarization framework inspired by the Mixture of Experts (MoE) paradigm, which operates as an inference-time algorithm without requiring any form of fine-tuning. Our approach integrates multiple VideoLLMs to generate comprehensive and coherent textual summaries. It effectively combines visual and audio content, provides detailed background descriptions, and excels at identifying keyframes, which enables more semantically meaningful retrieval compared to traditional computer vision approaches that rely solely on visual information, all without the need for additional fine-tuning. Moreover, the resulting summaries enhance performance in downstream tasks such as summary video generation, either through keyframe selection or in combination with text-to-image models. Our language-driven approach offers a semantically rich alternative to conventional methods and provides flexibility to incorporate newer VideoLLMs, enhancing adaptability and performance in video summarization tasks.
\begin{figure}[H]
    \includegraphics[width=0.9\linewidth]{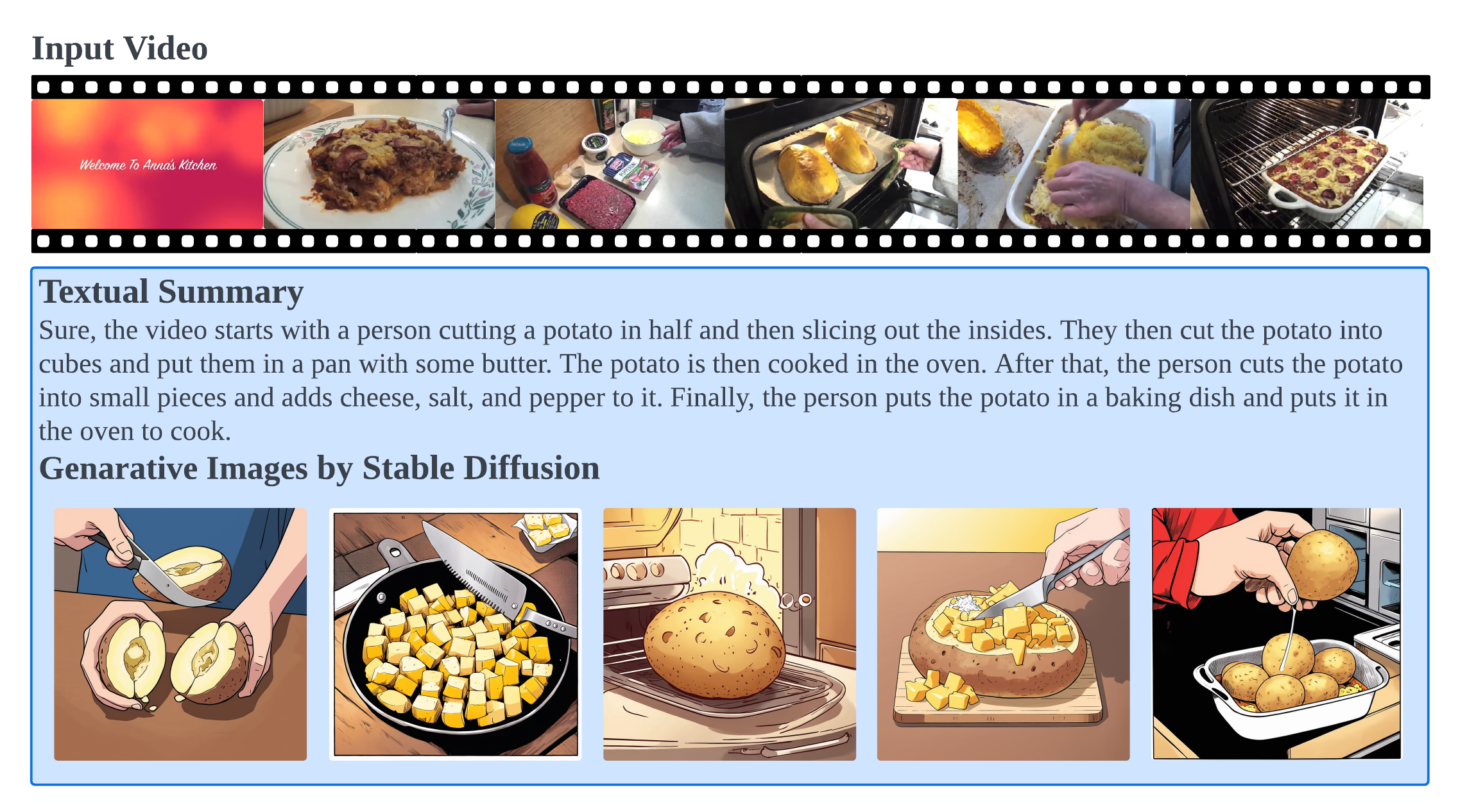}
    \caption{Visualization of two of our extended applications on HowTo100M \citep{miech2019howto100m}. The ``Input Video'' refers to the keyframes selected from the original video based on our textual summary, and pairing with our textual summary, we can simulate \textbf{visual manual generation}. The images generated by Stable Diffusion 3 \citep{esser2024scaling} simulate \textbf{privacy-preserving content generation}.}
    \label{fig:HowTo100M}
\end{figure}
\end{abstract}

\section{Introduction}
In recent day, the proliferation of video content across various platforms has led to an overwhelming amount of information, making it challenging for users to efficiently access and digest the key information. As a result, video summarization has emerged as a crucial task, enabling the efficient extraction of key segments from lengthy videos. The goal of video summarization is to condense extensive video content into textual summaries, short video clips, or a collection of representative images. Given the vast amount of video data generated daily, effective summarization not only enhances user experience by reducing the time required to access essential information but also supports efficient content management and retrieval across platforms. Additionally, video summarization has significant applications in areas such as surveillance, education, entertainment, and multimedia indexing, making it a vital tool for navigating and leveraging the vast expanse of video data available today.

The success of Visual Language Models (VLMs) \citep{liu2024visual, wang2023cogvlm, alayrac2022flamingo} has paved the way for the development of Video LLMs, such as VideoLLaMA \citep{zhang-etal-2023-video}, VideoChat \citep{li2023videochat}, and VideoLLaVA \citep{lin2023video}. These VideoLLMs leverage human-annotated data for instruction tuning, and they propose different methods to align video features with the LLM feature space. Each model exhibits distinct strengths: for example, PG-Video-LLaVA \citep{munasinghe2023PGVideoLLaVA} demonstrates pixel-grounded capabilities for capturing detailed scenes, Video-LLaMA adopts a multi-branch cross-modal framework that incorporates audio information in addition to video content, and LLaMA-VID \citep{li2023llamavid} excels in capturing background scene details. However, despite these strengths, existing VideoLLMs have inherent shortcomings and lack coherent methods to address them. For instance, Video-LLaVA and LLaMA-VID are unable to retrieve audio signals, while Video-LLaMA lacks the grounding abilities required for retrieving fine-grained details. Additionally, LLMs within these VideoLLMs often suffer from hallucination issues. To overcome these limitations, previous approaches typically resort to fine-tuning or retraining models, which can be computationally expensive. Our observation, however, suggests that the limitations of one VideoLLM can often be mitigated by the strengths of another. This leads us to ask: \textbf{\textit{What if we could utilize existing VideoLLMs collaboratively, instead of resorting to costly fine-tuning or retraining of a new model?}}

In this work, we draw inspiration from the Mixture of Experts (MoE) \citep{shen2023mixture, lin2024moe} paradigm, which is designed to enhance performance in processing large and complex tasks by leveraging multiple expert sub-models. Specifically, our approach employs multiple VideoLLMs for video summarization, integrating the concept of LLM cooperation to combine the outputs from these video ``experts'' through our proposed inference-time algorithm. This method allows us to address the limitations of individual VideoLLMs by compensating with the strengths of other expert VideoLLMs. Furthermore, since our framework does not require fine-tuning or retraining, it can seamlessly adapt to incorporate new or updated VideoLLMs as additional expert models.

Overall, we propose a novel video summarization method that follows a unique path of language-based semantic understanding. By proposing an inference-time algorithm, we can generate comprehensive textual summaries that capture not only visual content but also audio information, providing detailed descriptions of background scenes to offer users a more holistic view of the original videos. Additionally, with our comprehensive textual summaries, we can perform various downstream video summarization tasks, such as identifying keyframes and generating images and videos, thereby surpassing the capabilities of existing VideoLLMs.

Our main contributions can be summarized as follows:
\begin{itemize}
\item We propose an inference-time algorithm that leverages the capabilities of LLMs to combine the output summaries of multiple VideoLLMs into a single, coherent, and unbiased summary. This approach provides more detailed and comprehensive information, enhancing the overall quality of video summarization. 
\item Additionally, our comprehensive and coherent summaries enhance keyframe retrieval with a simple keyframe selection algorithm, surpassing the performance of existing approaches. 
\item Our proposed method is both flexible and general. The components of our framework can be easily replaced with more powerful models. Moreover, it is general enough to support extended video applications that can leverage our intermediate outputs, such as textual summaries and keyframes. 
\end{itemize}

\section{Related Work}
\paragraph{Large Language Models.} 
Large Language Models (LLMs) have revolutionized the field of natural language processing (NLP) and artificial general intelligence (AGI) with their exceptional capabilities in language generation, in-context learning, and reasoning. The historical evolution of these models began with foundational architectures such as BERT \citep{devlin2018bert}, GPT-2 \citep{radford2019language}, and T5 \citep{2020t5}, which set the stage for subsequent advancements. The introduction of GPT-3 \citep{brown2020language}, with its 175 billion parameters, marked a significant breakthrough, showcasing remarkable performance across a wide spectrum of language tasks. This progress spurred the development of an array of other influential LLMs, including Megatron-Turing NLG \citep{smith2022using}, Chinchilla \citep{hoffmann2022training}, PaLM \citep{chowdhery2023palm}, OPT \citep{zhang2022opt}, BLOOM \citep{le2023bloom}, LLaMA \citep{touvron2023llama}, MOSS \citep{Sun2024MOSS}, and GLM \citep{zeng2022glm}. These models, characterized by their scale and open-source availability, have become invaluable for both training large models and fine-tuning them for specific applications. 

\paragraph{Visual Language Models.} 
With the emergence of LLMs, recent works \citep{liu2024visual, wang2023cogvlm, alayrac2022flamingo} have increasingly explored their use in processing visual inputs, giving rise to Visual Language Models (VLMs). The central idea behind this line of work is to align visual features with the textual features of LLMs by utilizing a common framework. This framework typically involves a pretrained visual encoder to extract visual features, a projection layer to map these visual representations into the text latent space of LLMs, and the pretrained LLM to generate responses, thereby enabling the powerful capabilities of LLMs to be applied to vision tasks. Video-based Large Language Models (VideoLLMs) extend the capabilities of VLMs by incorporating temporal and/or audio features, allowing for richer video-language understanding through human-video dialogue interactions. For instance, methods such as VideoChatGPT \citep{Maaz2023VideoChatGPT} and Valley \citep{luo2023valley} use pooling over visual tokens to obtain compact visual representations. VideoChat \citep{li2023videochat} employs pretrained video foundation models and Q-Former \citep{zhang2024vision} from BLIP-2 \citep{li2022blip} to aggregate video representations. Video-LLaMA \citep{zhang-etal-2023-video} introduces a Video Q-Former and an Audio Q-Former for multimodal video comprehension. Furthermore, MovieChat \citep{song2024moviechat} proposes an advanced memory management mechanism for reasoning over extended video content.

\paragraph{LLM Evaluator.} The field of Natural Language Generation (NLG) evaluation has evolved considerably over the years, launching from traditional metrics to more advanced methodologies, particularly with the advent of LLMs. Early metrics, such as ROUGE \citep{lin2004rouge} and BLEU \citep{papineni2002bleu}, have been foundational in assessing the quality of generated text by comparing it to reference texts based on n-gram overlap. However, these methods have limitations in capturing deeper semantic nuances. To address this, embedding-based metrics like BERTScore \citep{zhang2019bertscore} were introduced, measuring the semantic similarity between texts using word and sentence embeddings. With the rise of LLMs, evaluation methods have further advanced. LLM-based evaluators, such as GPTScore \citep{fu2023gptscore}, G-Eval \citep{liu2023g}, and UniEval \citep{zhong2022towards}, leverage the comprehensive understanding and generation capabilities of LLMs to provide deeper insights into NLG quality. Recognizing the inherent limitations of these early approaches, subsequent studies concentrated on enhancing factual accuracy \citep{min2023factscore}, ensuring interpretability \citep{lu-etal-2024-error}, reducing position bias \citep{wang2023large}, and aligning evaluation more closely with human judgment standards \citep{liu2023calibrating}. These efforts represent a significant shift toward more robust and human-aligned evaluation methods in NLG.
\begin{figure}
\hskip -1.8em
\includegraphics[width=1.1\textwidth]{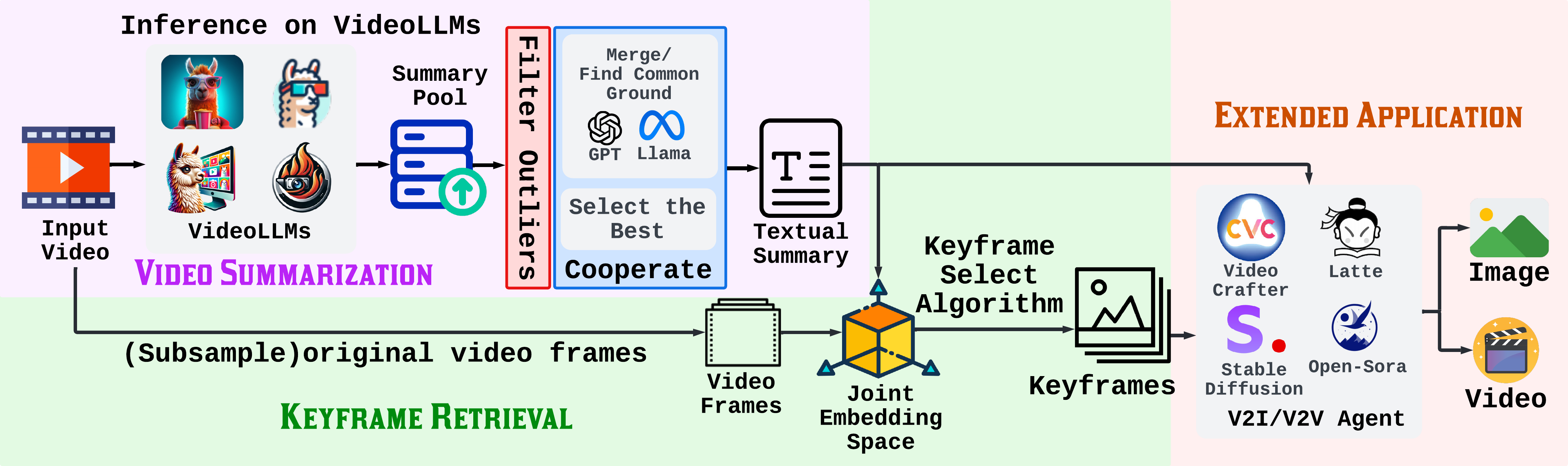}
\caption{An overview of our framework. Our approach consists of three main modules: \textbf{(1) Video Summarization}, which constructs coherent textual summaries by leveraging multiple existing VideoLLMs and our proposed inference-time algorithm; \textbf{(2) Keyframe Retrieval}, which identifies key moments based on our textual summary using a simple keyframe selection algorithm; and \textbf{(3) Extended Applications}, which utilize our informative textual summaries and keyframes to address real-world tasks beyond traditional video summarization.}
\label{fig:framework}
\end{figure}

\section{Methodology}
Our holistic video summarization framework, illustrated in Figure \ref{fig:framework}, is composed of three key modules. In Section \ref{sec:video-summarize}, we introduce two components that utilize VideoLLM ``experts'' to produce textual summaries within the video summarization module. In Section \ref{sec:keyframe-retrieval}, we present our keyframe retrieval module, which details how video frames and textual summaries are projected into a joint embedding space to identify relevant keyframes. Finally, in Section \ref{sec:application}, we explore the extended applications of our framework, demonstrating how the textual summaries and corresponding keyframes can be used for real-world applications.

\subsection{Video Summarization} \label{sec:video-summarize}
We perform inference on the given input video using multiple VideoLLMs. To fully leverage the capabilities of these models, we design and employ prompts specifically tailored to the architecture of each VideoLLM. This approach results in four unique summaries, each capturing different aspects of the input video and reflecting the strengths of each model.

\subsubsection{Denoise-and-Cooperate} 
There are two primary challenges in utilizing the generated summaries from these VideoLLMs. First, each VideoLLM exhibits varying degrees of the ``hallucination'' issue, which can mislead users and make us difficult to identify the inaccuracies specific to each model. Second, effectively integrating and combining the ``strengths'' of each model from the resulting summaries is a complex task. To address these challenges, we propose the following strategies:

\textbf{Filter Outliers.} We propose two outlier filtering strategies to remove the summaries that deviate from the others, that is, from the four distinct summaries generated in the previous step, we identify and exclude the summary that exhibits the lowest similarity to the other three, considering it an outlier. For the first strategy, we reference the scoring method from Open-Sora\footnote{\href{https://github.com/hpcaitech/Open-Sora/tree/main/tools/scoring}{https://github.com/hpcaitech/Open-Sora/tree/main/tools/scoring} (last accessed: 2024/09)} to evaluate the summaries generated by each VideoLLM. By calculating the matching score between each summary and the middle frame of the video, we identify and remove the summary with the lowest score. As for the second strategy, we aim to enhance the video-text alignment between the generated summaries and the input video, our implementation is outlined in Algorithm \ref{algo:avg_clip}. This involves calculating the average CLIP score across the summaries and discarding the one with the lowest score.

\textbf{Cooperate.} After filtering outliers, we leverage the capabilities of state-of-the-art LLMs, to combine the remaining summaries into a single coherent paragraph. We propose three distinct strategies for this synthesis: Merge, Find Common Ground, and Select.

\begin{itemize}[leftmargin=*]
\item Merge: This strategy integrates all information from the VideoLLM summaries into a comprehensive single summary, capturing the full spectrum of details provided by each model. The resulting summary aims to be inclusive and detailed.
\item Find Common Ground: This approach focuses on extracting and consolidating only the common elements across all VideoLLM summaries. The process produces a coherent summary that emphasizes the most consistent and reliable information, potentially reducing noise and inconsistencies.
\item Select: This strategy chooses the summary that achieves the highest score based on our evaluation metric in the outlier filtering stage. We find this approach particularly effective for certain video types, such as instructional videos in datasets like HowTo100M \citep{miech2019howto100m}. 
\end{itemize}

These strategies provide flexibility in addressing various video content types, allowing for adaptability in the fusion process. The choice of strategy can be tailored to the specific needs of the task or the nature of the video content being summarized.

\begin{algorithm}
\caption{Calculate Average CLIP Score and Remove Minimum}
\begin{algorithmic}[1]
    \Require{\text{summary }$s_i \in \mathcal{S} $, \text{video\_frames} $f_i \in \mathcal{F}$}  
    \State $\text{Initialize empty average CLIP score list }\bar{\mathcal{C}}$
    \For{$s_i \in \mathcal{S}$}
        \State $\text{Calculate average CLIP score of $s_i$ with respect to $\mathcal{F}$: } \Bar{c}= \frac{1}{|\mathcal{F}|} \sum_{f_i \in \mathcal{F}} \mathrm{CLIP}(s_i, f_i)$
        \State $\text{Store $\bar{c}$ to $\bar{\mathcal{C}}$}$
    \EndFor 
    \State $\text{Locate the index $j$ of the lowest score in $\bar{\mathcal{C}}$}$
    \State $\text{Remove } s_j \text{ from } \mathcal{S}.$
 \end{algorithmic}
 \label{algo:avg_clip}
\end{algorithm}

\subsection{Keyframe Retrieval} \label{sec:keyframe-retrieval}
After obtaining our coherent summary from the Video Summarization module, previous methods either prompt the VideoLLM to generate short segments most relevant to the summary \citep{qianmomentor, huang2024vtimellm}, which is an area where current VideoLLMs often underperform, or training a model specifically to encode visual and textual features \citep{lin2023univtg, moon2023correlation}. The latter approach often employs a sliding window technique to capture and align temporal information, enabling the accurate identification and retrieval of relevant video segments that correspond to the summary. However, this method is computationally expensive and can sometimes result in redundant information.

Given that our textual summary is highly informative, we propose an alternative approach that avoids the need for training a new model. Instead, we utilize a fixed joint embedding space, combined with a similarity metric, to guide the keyframe retrieval. Specifically, we encode the input video frames at two-second intervals, following a sampling technique inspired by Moment-DETR \citep{lei2021detecting}, alongside our textual summary. Both the text and video frames are encoded using CLIP \citep{radford2021learning}. We then calculate the cosine similarity between the text embeddings (whole summary) and the individual frame embeddings, sorting the similarity scores in descending order to identify the top-$k$ video frames as keyframes.

\subsection{Extended Applications} \label{sec:application}
Our method extends beyond existing video summarization, offering practical real-world applications that leverage both our coherent textual summary (from Section \ref{sec:video-summarize}) and retrieved keyframes (from Section \ref{sec:keyframe-retrieval}). These include visual manual generation for instructional videos, aiding product manufacturers in creating efficient user guides, and privacy-preserving content generation, which produces short video clips and representative images that capture the essence of the original video without revealing sensitive content. These applications demonstrate our method's versatility, addressing challenges in content creation, information dissemination, and privacy protection across various domains, thus surpassing the capabilities of existing VideoLLMs.
\section{Experiments}
\subsection{Experimental Setup} \label{sec:exp_setup}
\paragraph{\uline{Dataset.}} We evaluate our approach on four well-established datasets: QVHighlights \citep{lei2021detecting}, TACoS \citep{regneri2013grounding}, Charades-STA \citep{gao2017tall} and DiDeMo \citep{anne2017localizing}. These datasets span diverse video domains, including sports, product reviews, cooking scenarios, and household activities, etc., providing a comprehensive foundation for assessing our method's performance. Table \ref{tab:dataset} summarizes the key characteristics of each dataset.

\begin{table}[]
\centering
\small
\caption{Statistics of datasets used in our evaluation.}
\label{tab:dataset}
\begin{tabular}{lcccp{5cm}}
\toprule
Dataset & Videos & Video-Qeury Pairs & Avg. Video Len (sec) & Video Types \\
\midrule
QVHighlights & 10,148 & 10,310 & 150 & Diverse (daily, travel, news, etc.) \\
TACoS & 127 & 18,818 & 287 & Cooking \\
Charades-STA & 9,848 & 18,131 & 30 & Indoor activities \\
DiDeMo & 10,464 & 40,543 & 30 & Diverse (from Flickr) \\
\bottomrule
\end{tabular}
\end{table}

\paragraph{\uline{Implementation Details.}} In our experiments, we employ four VideoLLM ``experts'': Video-LLaMA \citep{zhang-etal-2023-video}, Video-LLaVA \citep{lin2023video}, PG-Video-LLaVA \citep{munasinghe2023PGVideoLLaVA}, and LLaMA-VID \citep{li2023llamavid}, which serve as both components of our approach and individual baselines. We obtain summaries from each expert, use average CLIP scores to remove outliers, and apply our \textbf{Find Common Ground} strategy with Llama-3-8B-Instruct\footnote{\href{https://huggingface.co/meta-llama/Meta-Llama-3-8B-Instruct}{https://huggingface.co/meta-llama/Meta-Llama-3-8B-Instruct} (last accessed: 2024/09)} to synthesize the final coherent summary. For keyframe retrieval task, we encode video frames (sampled at two-second intervals) and our generated textual summary into CLIP \citep{radford2021learning} embedding space before calculating similarity metrics.

\subsection{Experimental Results}
\paragraph{\uline{Textual Video Summarization.}} To evaluate the quality of our textual summaries and their alignment with ground truth, we employ G-Eval \citep{liu2023g}, which we utilize GPT-4-Turbo\footnote{\href{https://openai.com/index/gpt-4/}{https://openai.com/index/gpt-4/} (last accessed: 2024/09)} as the LLM backbone. This method evaluates summaries across seven dimensions: aspect coverage, coherence, faithfulness, fluency, relevance, sentiment consistency, and specificity. Importantly, G-Eval not only assesses video-text alignment through the relevance score but also provides insights into potential human preferences through the remaining metric scores. The results, presented in Table \ref{tab:textual_summary}, demonstrate that our generated summaries consistently outperform all baseline methods. Our approach achieves superior scores in both video-text alignment and across all aspects that typically correlate with human preference. This comprehensive evaluation underscores the effectiveness of our method in producing high-quality, relevant, and potentially more appealing summaries compared to existing approaches. We also present qualitative results comparing our textual summaries with those of baseline models in Figure \ref{fig:Textual_summary_QV} (More qualitative results on different datasets are provided in the Appendix, cf. Figure \ref{fig:textual_charades} and \ref{fig:textual_tacos}). While most summaries generated by our baselines capture the essential content, but our approach captures a broader spectrum of information from the given video, providing a more complete and nuanced representation of the content. Also, our method demonstrates potential as an automatic (re-)annotation tool. In cases where ground truth summaries may be inaccurate, as shown in our qualitative results, our framework can serve as a valuable means to verify and potentially correct existing annotations. This capability highlights an additional extensibility of our approach, offering a robust mechanism for enhancing the quality and reliability of video annotation datasets.

\begin{figure}
\hskip -5em
\includegraphics[width=1.25\textwidth]{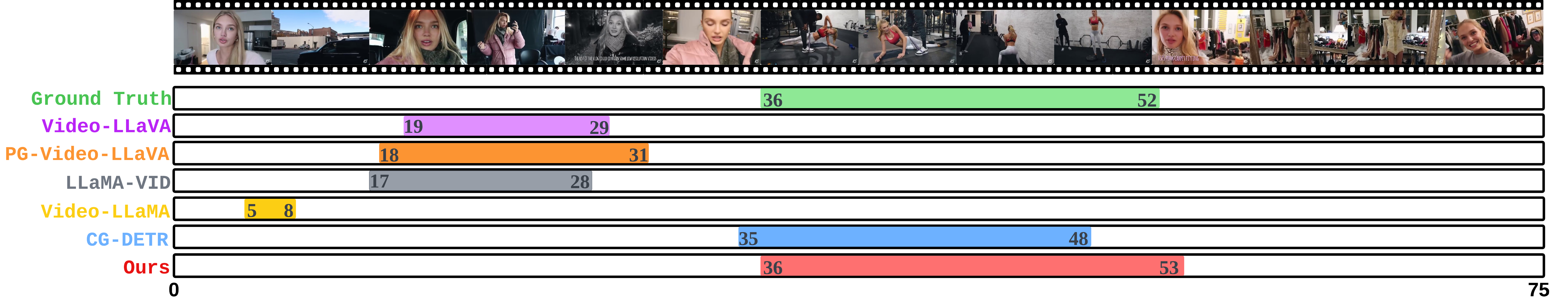}
\caption{Visualization of prediction results comparison on QVHighlights \citep{lei2021detecting}. The ground truth keyframes are shown at the top as the input video, and the prediction unit is in seconds.}
\label{fig:QVHighlight}
\end{figure}

\paragraph{\uline{Visual Keyframe Retrieval.}} Following the evaluation metrics in TVR \citep{lei2020tvr} and Tall \citep{regneri2013grounding}, we compute the mean Intersection over Union (mIoU) and Recall@1 with IoU thresholds of 0.5, and 0.7. In addition to individual VideoLLMs as prompt-based baselines, we also include CG-DETR \citep{moon2023correlation} as the query-based baseline. The results, presented in Table \ref{tab:keyframe_retrieval}, demonstrate our approach's effectiveness. We outperform all baselines on the Charades-STA, TACoS and DiDeMo datasets, and surpass prompt-based baselines on QVHighlights. Notably, our method, without fine-tuning, achieves superior performance on Charades-STA, TACoS and DiDeMo compared to the fine-tuned CG-DETR. While CG-DETR shows better results on QVHighlights, it's important to consider that CG-DETR benefits from dataset-specific fine-tuning. In contrast, our method's strong performance across datasets in a zero-shot setting underscores its robust generalization capabilities. We also provide the qualitative comparison of our keyframe retrieval results against those of our baselines in Figure \ref{fig:QVHighlight}. The visual comparison clearly demonstrates that our selected keyframes achieve a significantly higher coverage rate of the ground truth compared to prompt-based baselines. Moreover, our approach shows superior performance even when compared to CG-DETR. These results visually reinforce the quantitative findings, highlighting our method's effectiveness in accurately identifying and retrieving key moments from videos.

\paragraph{\uline{Extended Applications.}} We demonstrate two extended applications of our framework on the HowTo100M \citep{miech2019howto100m} dataset, which primarily consists of instructional videos. Figure \ref{fig:HowTo100M} presents the qualitative results of these applications, and more results are provided in the Appendix (cf. Figure \ref{fig:HowTo100M_Appendix}).
For \textbf{visual manual generation}, our generated summary mimic the textual instruction, and the selected keyframes are the visual instructions. This combination of textual and visual elements effectively simulates the creation of visual manuals for instructional content. In the \textbf{privacy-preserving content generation}, we utilize Stable Diffusion 3 \citep{esser2024scaling} to generate images based on our textual summaries. The resulting images successfully interpret the content of the original videos without revealing sensitive information. These qualitative results illustrate the versatility of our framework in generating practical, real-world applications beyond standard video summarization tasks.

\begin{table}[]
\small
\caption{Quantitative evaluation of our generated textual video summary among various approaches with G-Eval \citep{liu2023g}. The best results are marked in \textbf{bold}.}
\begin{tabular}{lccccc}
\toprule
Dimension & OURS & Video-LLaVA & PG-Video-LLaVA & LLaMA-VID & Video-LLaMA \\
\midrule
aspect coverage & \textbf{2.77} & 1.31 & 1.85 & 1.97 & 1.72 \\
coherence & \textbf{3.35} & 1.56 & 2.12 & 2.76 & 1.83 \\
faithfulness & \textbf{2.14} & 1.31 & 1.63 & 1.65 & 1.49 \\
fluency & \textbf{3.31} & 1.66 & 2.29 & 2.89 & 2.01 \\
relevance & \textbf{2.59} & 1.5 & 1.66 & 1.96 & 1.42 \\
sentiment consistency & \textbf{1.92} & 1.23 & 1.38 & 1.6 & 1.31 \\
specificity & \textbf{3.22} & 1.41 & 2.12 & 2.44 & 1.97 \\
\toprule
\end{tabular}
\label{tab:textual_summary}
\end{table}

\begin{figure}
\centering
\includegraphics[width=1.1\textwidth]{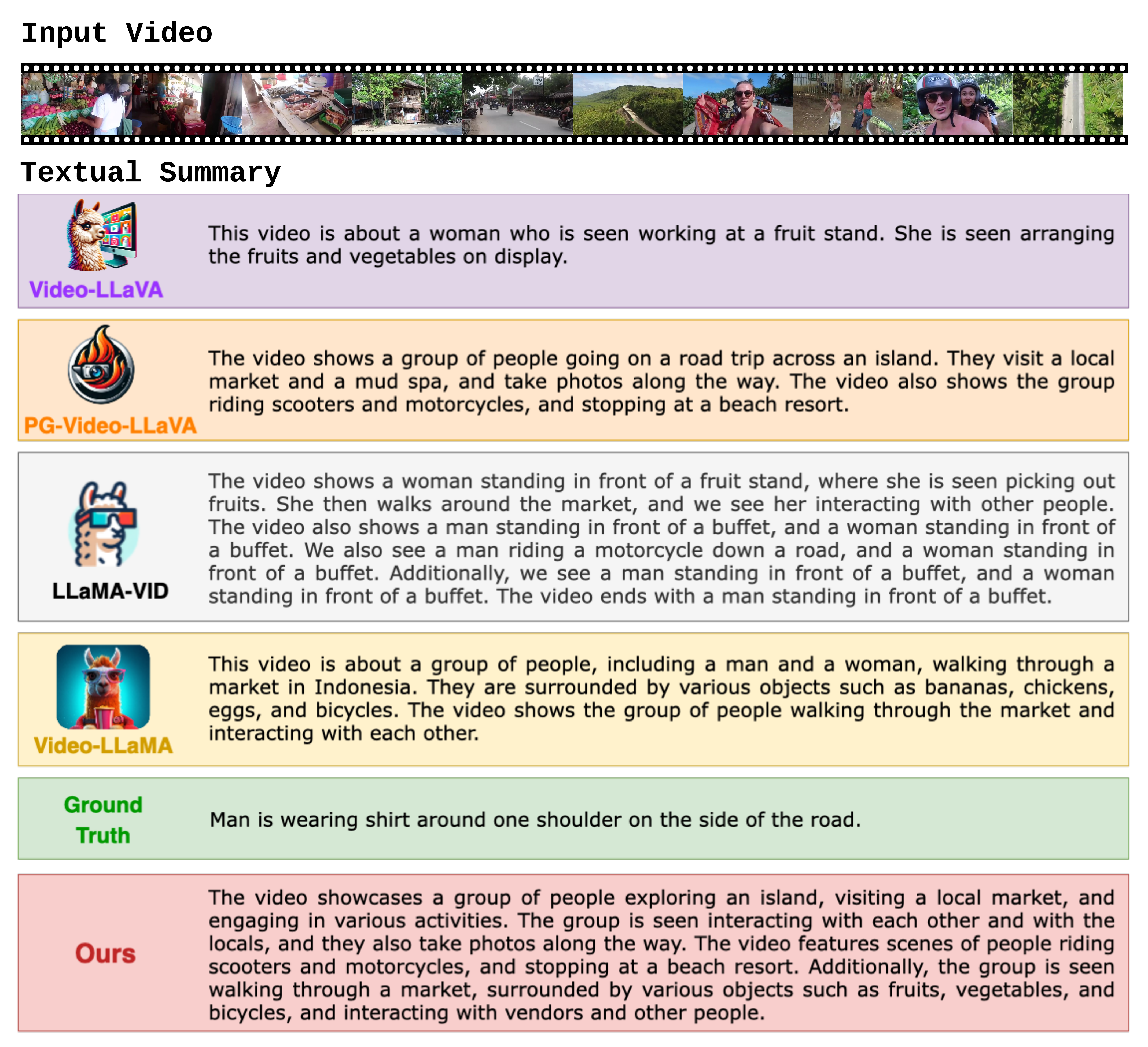}
\caption{Visualization of textual video summaries generated by individual VideoLLMs and our proposed collaboration approach. Keyframes are displayed at the top as input video. Additionally, we provide the ground truth summary for reference.
}
\label{fig:Textual_summary_QV}
\end{figure}

\begin{table}[h]
\vskip 5em
\caption{Quantitative evaluation of our keyframe retrieval prediction among prompt-based and query-based approaches. The best results are marked in \textbf{bold}, and the second-best results are \uline{underlined}.}
\adjustbox{max width=1.1\textwidth}{
\begin{tabular}{lcccccccccccc}
\toprule
\multirow{2}{*}{Methods} & \multicolumn{3}{c}{Charades-STA} & \multicolumn{3}{c}{QVHighlights} & \multicolumn{3}{c}{TACoS} & \multicolumn{3}{c}{DiDeMo} \\
\cmidrule(lr){2-4} \cmidrule(lr){5-7}  \cmidrule(lr){8-10} \cmidrule(lr){11-13}
& mIoU & R@0.5 & R@0.7 & mIoU & R@0.5 & R@0.7 & mIoU & R@0.5 & R@0.7 & mIoU & R@0.5 & R@0.7 \\
\midrule
\multicolumn{13}{l}{\emph{prompt-based}} \\
Video-LLaVA & 0.68 & 0.3 & 0.07 & 9 & 3 & 1.2 & 10.09 & 0.37 & 0 & 6.83 & 2.49 & 0 \\
Video-LLaMA & 5.54 & 1.64 & 0.48 & 5.1 & 1.6 & 0 & 0.13 & 0 & 0 & 6.01 & 0.6 & 0 \\
LLAMA-VID & 20 & 13.79 & {\ul 6.73} & 13.8 & 9.3 & 3.6 & 8.23 & 0.34 & 0 & 17.28 & 9.8 & 3.7 \\
PG-Video-LLaVA & 2.57 & 1.32 & 0.57 & 10.1 & 4.4 & 1.3 & 5.7 & 0 & 0 & 7.63 & 1.97 & 0.19 \\
\midrule
\multicolumn{13}{l}{\emph{query-based}} \\
CG-DETR & {\ul 26.33} & {\ul 14.86} & 6.2 & \textbf{53.62} & \textbf{54.47} & \textbf{42.29} & {\ul 31.22} & {\ul 8.46} & {\ul 7.35} & {\ul 17.69} & {\ul 10.24} & {\ul 3.25} \\
\midrule
Ours & \textbf{35.72} & \textbf{27.95} & \textbf{13.88} & {\ul 24.08} & {\ul 15.33} & {\ul 10.29} & \textbf{94.17} & \textbf{96.93} & \textbf{96.93} & \textbf{21.96} & \textbf{10.56} & \textbf{3.92} \\
\bottomrule
\end{tabular}
}
\label{tab:keyframe_retrieval}
\vskip -1em
\end{table}

\subsection{Ablation Studies}
For the experiments in the following studies, the experimental setup follows our main setting in Section \ref{sec:exp_setup}, and we focus on the keyframe retrieval task evaluated on three datasets: QVHighlights, Charades-STA, and TACoS, and metrics: mIoU, R@0.3, R@0.5, and R@0.7, unless otherwise specified.

\paragraph{\uline{Effect of filtering outliers.}} To assess the impact of our ``Filter Outliers'' component, we compare our framework's performance with and without this feature. In both scenarios, we utilize GPT-4-Turbo to synthesize summaries from individual VideoLLMs using the ``Find Common Ground'' strategy. The key difference lies in the input to this fusion process: with outlier filtering, we exclude the detected outlier, while without it, all four summaries are included. As demonstrated in Table \ref{tab:ablation_filter_outliers}, the inclusion of outlier filtering led to a significant improvement in performance, enhancing both mean Intersection over Union (mIoU) and Recall metrics by at least 2\%. This consistent improvement across metrics underscores the effectiveness of our outlier filtering approach in refining the quality of the final summary.

\paragraph{\uline{Effect of different cooperation strategies with different LLMs.}} We examine the impact of different cooperation strategies and LLMs on our framework's performance. We compare two cooperation strategies, Merge and Find Common Ground, implemented with two distinct LLMs: the open-source Llama-3-8b-Instruct and the closed-source GPT-4-Turbo, and the prompt template we apply is provided in the Appendix (cf. \ref{sec:merging_prompt}). We present the results in Table \ref{tab:ablation_combine_LLM}. Our analysis reveals that the choice of LLM and cooperation strategy has only marginal effects on the overall performance. However, all combinations demonstrate substantial improvements over utilizing only the individual VideoLLM summaries, as shown in Table \ref{tab:keyframe_retrieval}. Our results strongly suggest that our method of combining and refining summaries from multiple VideoLLMs produces more comprehensive and accurate textual representations, which in turn lead to improved keyframe selection.

\begin{table}[]
\centering
\caption{Ablation study of the effect of filtering outliers.}
\adjustbox{max width=1.1\textwidth}{
\begin{tabular}{lcccccccccccc}
\toprule
\multirow{2}{*}{} & \multicolumn{4}{c}{QVHighlights} & \multicolumn{4}{c}{Charades-STA} & \multicolumn{4}{c}{TACoS} \\
\cmidrule(lr){2-5} \cmidrule(lr){6-9}  \cmidrule(lr){10-13}
 & mIoU & \multicolumn{1}{c}{R@0.3} & \multicolumn{1}{l}{R@0.5} & \multicolumn{1}{l}{R@0.7} & mIoU & \multicolumn{1}{c}{R@0.3} & \multicolumn{1}{l}{R@0.5} & \multicolumn{1}{l}{R@0.7} & mIoU & \multicolumn{1}{c}{R@0.3} & \multicolumn{1}{l}{R@0.5} & \multicolumn{1}{l}{R@0.7} \\
\midrule
w Filter outliers & \multicolumn{1}{r}{23.93} & 21.93 & 14.88 & 10.29 & \multicolumn{1}{r}{35.16} & 36.1 & 27.03 & 13.94 & \multicolumn{1}{r}{94.37} & 97.04 & 97.04 & 97.04 \\
w/o Filter outliers & \multicolumn{1}{r}{21.92} & 19.53 & 12.9 & 7.35 & \multicolumn{1}{r}{N/A} & N/A & N/A & N/A & \multicolumn{1}{r}{92.27} & 95.89 & 95.89 & 95.89 \\
\bottomrule
\end{tabular}
}
\label{tab:ablation_filter_outliers}
\end{table}

\begin{table}[]
\centering
\caption{\textbf{Ablation study of effects of different cooperation strategies utilizing different LLMs.} In the "LLM" column, ``GPT'' represents ``GPT-4-Turbo'', and ``LLaMA'' denotes ``LLaMA-3-8b-Instruct''. In the ``Cooperate Strategy'' column, ``CG'' refers to the ``Find Common Ground'' strategy, while ``M'' stands for the ``Merge'' strategy. The best results are marked in \textbf{bold}.}
\adjustbox{max width=1.1\textwidth}{
\begin{tabular}{clrrrrrrrrrrrr}
\toprule
\multirow{2}{*}{LLM} & \multicolumn{1}{c}{\multirow{2}{*}{\begin{tabular}[c]{@{}c@{}}Cooperate \\ Strategy\end{tabular}}} & \multicolumn{4}{c}{QVHighlights} & \multicolumn{4}{c}{Charades-STA} & \multicolumn{4}{c}{TACoS} \\
\cmidrule(lr){3-6} \cmidrule(lr){7-10}  \cmidrule(lr){11-14}
 & \multicolumn{1}{c}{} & \multicolumn{1}{c}{mIoU} & \multicolumn{1}{c}{R@0.3} & \multicolumn{1}{l}{R@0.5} & \multicolumn{1}{l}{R@0.7} & \multicolumn{1}{c}{mIoU} & \multicolumn{1}{c}{R@0.3} & \multicolumn{1}{l}{R@0.5} & \multicolumn{1}{l}{R@0.7} & \multicolumn{1}{c}{mIoU} & \multicolumn{1}{c}{R@0.3} & \multicolumn{1}{l}{R@0.5} & \multicolumn{1}{l}{R@0.7} \\
\midrule
\multirow{2}{*}{GPT} & CG & 23.93 & 21.93 & 14.88 & \textbf{10.29} & 35.16 & 36.1 & 27.03 & 13.94 & 94.37 & 97.04 & 97.04 & 97.04 \\
 & M & 23.45 & 21.23 & 14.13 & 9.81 & \textbf{35.81} & 37.3 & 27.82 & \textbf{14} & \textbf{94.4} & \textbf{97.06} & \textbf{97.06} & \textbf{97.06} \\
\multirow{2}{*}{LLaMA} & CG & \textbf{24.08} & \textbf{22.31} & \textbf{15.33} & \textbf{10.29} & 35.41 & 37.33 & 27.27 & 13.72 & 94.05 & 96.86 & 96.86 & 96.31 \\
 & M & 23.1 & 20.84 & 13.92 & 9.43 & 35.72 & \textbf{37.76} & \textbf{27.95} & 13.88 & 94.18 & 96.93 & 96.93 & 96.93 \\
 \bottomrule
\end{tabular}
}
\label{tab:ablation_combine_LLM}
\end{table}

\paragraph{\uline{Effect of audio information.}} To assess the influence of audio information, we conduct experiments with and without audio input, noting that some VideoLLMs, such as Video-LLaMA, incorporate audio information, others like Video-LLaVA and LLaMA-VID do not include this modality in their frameworks. For Video-LLaMA, we remove the audio branch to simulate scenarios without audio information. In the case of PG-Video-LLaVA, we deactivate the audio branch in our default setting. The results, presented in Table \ref{tab:ablation_audio}, demonstrate the significant contribution of audio information to the quality of video summaries. Including audio led to a 5-10\% improvement in downstream keyframe retrieval performance.

\begin{table}[H]
\centering
\caption{\textbf{Ablation study of the impact of audio information.} We remove the audio branch of Video-LLaMA \citep{zhang-etal-2023-video} to simulate the case of ``w/o audio''.}
\begin{tabular}{lcrrrcrrr}
\toprule
\multirow{2}{*}{} & \multicolumn{4}{c}{QVHighlights} & \multicolumn{4}{c}{Charades-STA} \\
\cmidrule(lr){2-5} \cmidrule(lr){6-9}
 & mIoU & \multicolumn{1}{c}{R@0.3} & \multicolumn{1}{l}{R@0.5} & \multicolumn{1}{l}{R@0.7} & mIoU & \multicolumn{1}{c}{R@0.3} & \multicolumn{1}{l}{R@0.5} & \multicolumn{1}{l}{R@0.7} \\
\midrule
w audio & \multicolumn{1}{r}{24.08} & 22.31 & 15.33 & 10.29 & \multicolumn{1}{r}{35.16} & 36.1 & 27.03 & 13.94 \\
w/o audio & \multicolumn{1}{r}{19.45} & 17.28 & 10.73 & 7.1 & \multicolumn{1}{r}{27.75} & 22.24 & 10.58 & 2.5 \\
 \bottomrule
\end{tabular}
\label{tab:ablation_audio}
\end{table}

\section{Conclusion}
We propose a holistic video summarization framework that leverages multiple VideoLLMs to generate comprehensive textual summaries that capture the detail of the given video without fine-tuning. Our extensive experiments demonstrate the effectiveness of our method in downstream tasks like keyframe retrieval and extended applications such as visual manual generation and privacy-preserving content creation. Our framework's adaptability allows for easy integration of more advanced models, ensuring its relevance as the field progresses. By establishing a foundation for integrating visual and linguistic information, our approach paves the way for more sophisticated multimedia analysis tools. We anticipate that this framework will catalyze advancements in video understanding and natural language processing, leading to more intuitive and powerful systems across various domains.

\bibliography{main}

\begin{thebibliography}{49}
\providecommand{\natexlab}[1]{#1}
\providecommand{\url}[1]{\texttt{#1}}
\expandafter\ifx\csname urlstyle\endcsname\relax
  \providecommand{\doi}[1]{doi: #1}\else
  \providecommand{\doi}{doi: \begingroup \urlstyle{rm}\Url}\fi

\bibitem[Alayrac et~al.(2022)Alayrac, Donahue, Luc, Miech, Barr, Hasson, Lenc, Mensch, Millican, Reynolds, et~al.]{alayrac2022flamingo}
Jean-Baptiste Alayrac, Jeff Donahue, Pauline Luc, Antoine Miech, Iain Barr, Yana Hasson, Karel Lenc, Arthur Mensch, Katherine Millican, Malcolm Reynolds, et~al.
\newblock Flamingo: a visual language model for few-shot learning.
\newblock In \emph{NeurIPS}, 2022.

\bibitem[Anne~Hendricks et~al.(2017)Anne~Hendricks, Wang, Shechtman, Sivic, Darrell, and Russell]{anne2017localizing}
Lisa Anne~Hendricks, Oliver Wang, Eli Shechtman, Josef Sivic, Trevor Darrell, and Bryan Russell.
\newblock Localizing moments in video with natural language.
\newblock In \emph{ICCV}, 2017.

\bibitem[Brown(2020)]{brown2020language}
Tom~B Brown.
\newblock Language models are few-shot learners.
\newblock \emph{arXiv preprint ArXiv:2005.14165}, 2020.

\bibitem[Chowdhery et~al.(2023)Chowdhery, Narang, Devlin, Bosma, Mishra, Roberts, Barham, Chung, Sutton, Gehrmann, et~al.]{chowdhery2023palm}
Aakanksha Chowdhery, Sharan Narang, Jacob Devlin, Maarten Bosma, Gaurav Mishra, Adam Roberts, Paul Barham, Hyung~Won Chung, Charles Sutton, Sebastian Gehrmann, et~al.
\newblock Palm: Scaling language modeling with pathways.
\newblock \emph{JMLR}, 2023.

\bibitem[Devlin(2018)]{devlin2018bert}
Jacob Devlin.
\newblock Bert: Pre-training of deep bidirectional transformers for language understanding.
\newblock \emph{arXiv preprint arXiv:1810.04805}, 2018.

\bibitem[Esser et~al.(2024)Esser, Kulal, Blattmann, Entezari, M{\"u}ller, Saini, Levi, Lorenz, Sauer, Boesel, et~al.]{esser2024scaling}
Patrick Esser, Sumith Kulal, Andreas Blattmann, Rahim Entezari, Jonas M{\"u}ller, Harry Saini, Yam Levi, Dominik Lorenz, Axel Sauer, Frederic Boesel, et~al.
\newblock Scaling rectified flow transformers for high-resolution image synthesis.
\newblock In \emph{ICML}, 2024.

\bibitem[Fu et~al.(2023)Fu, Ng, Jiang, and Liu]{fu2023gptscore}
Jinlan Fu, See-Kiong Ng, Zhengbao Jiang, and Pengfei Liu.
\newblock Gptscore: Evaluate as you desire.
\newblock \emph{arXiv preprint arXiv:2302.04166}, 2023.

\bibitem[Gao et~al.(2017)Gao, Sun, Yang, and Nevatia]{gao2017tall}
Jiyang Gao, Chen Sun, Zhenheng Yang, and Ram Nevatia.
\newblock Tall: Temporal activity localization via language query.
\newblock In \emph{ICCV}, 2017.

\bibitem[Hoffmann et~al.(2022)Hoffmann, Borgeaud, Mensch, Buchatskaya, Cai, Rutherford, Casas, Hendricks, Welbl, Clark, et~al.]{hoffmann2022training}
Jordan Hoffmann, Sebastian Borgeaud, Arthur Mensch, Elena Buchatskaya, Trevor Cai, Eliza Rutherford, Diego de~Las Casas, Lisa~Anne Hendricks, Johannes Welbl, Aidan Clark, et~al.
\newblock Training compute-optimal large language models.
\newblock \emph{arXiv preprint arXiv:2203.15556}, 2022.

\bibitem[Huang et~al.(2024)Huang, Wang, Chen, Song, and Zhu]{huang2024vtimellm}
Bin Huang, Xin Wang, Hong Chen, Zihan Song, and Wenwu Zhu.
\newblock Vtimellm: Empower llm to grasp video moments.
\newblock In \emph{CVPR}, 2024.

\bibitem[Le~Scao et~al.(2023)Le~Scao, Fan, Akiki, Pavlick, Ili{\'c}, Hesslow, Castagn{\'e}, Luccioni, Yvon, Gall{\'e}, et~al.]{le2023bloom}
Teven Le~Scao, Angela Fan, Christopher Akiki, Ellie Pavlick, Suzana Ili{\'c}, Daniel Hesslow, Roman Castagn{\'e}, Alexandra~Sasha Luccioni, Fran{\c{c}}ois Yvon, Matthias Gall{\'e}, et~al.
\newblock Bloom: A 176b-parameter open-access multilingual language model.
\newblock \emph{arXiv preprint arXiv:2211.05100}, 2023.

\bibitem[Lei et~al.(2020)Lei, Yu, Berg, and Bansal]{lei2020tvr}
Jie Lei, Licheng Yu, Tamara~L Berg, and Mohit Bansal.
\newblock Tvr: A large-scale dataset for video-subtitle moment retrieval.
\newblock In \emph{ECCV}, 2020.

\bibitem[Lei et~al.(2021)Lei, Berg, and Bansal]{lei2021detecting}
Jie Lei, Tamara~L Berg, and Mohit Bansal.
\newblock Detecting moments and highlights in videos via natural language queries.
\newblock In \emph{NeurIPS}, 2021.

\bibitem[Li et~al.(2022)Li, Li, Xiong, and Hoi]{li2022blip}
Junnan Li, Dongxu Li, Caiming Xiong, and Steven Hoi.
\newblock Blip: Bootstrapping language-image pre-training for unified vision-language understanding and generation.
\newblock In \emph{ICML}, 2022.

\bibitem[Li et~al.(2023{\natexlab{a}})Li, He, Wang, Li, Wang, Luo, Wang, Wang, and Qiao]{li2023videochat}
KunChang Li, Yinan He, Yi~Wang, Yizhuo Li, Wenhai Wang, Ping Luo, Yali Wang, Limin Wang, and Yu~Qiao.
\newblock Videochat: Chat-centric video understanding.
\newblock \emph{arXiv preprint arXiv:2305.06355}, 2023{\natexlab{a}}.

\bibitem[Li et~al.(2023{\natexlab{b}})Li, Wang, and Jia]{li2023llamavid}
Yanwei Li, Chengyao Wang, and Jiaya Jia.
\newblock Llama-vid: An image is worth 2 tokens in large language models.
\newblock \emph{arXiv preprint arXiv:2311.17043}, 2023{\natexlab{b}}.

\bibitem[Lin et~al.(2023{\natexlab{a}})Lin, Zhu, Ye, Ning, Jin, and Yuan]{lin2023video}
Bin Lin, Bin Zhu, Yang Ye, Munan Ning, Peng Jin, and Li~Yuan.
\newblock Video-llava: Learning united visual representation by alignment before projection.
\newblock \emph{arXiv preprint arXiv:2311.10122}, 2023{\natexlab{a}}.

\bibitem[Lin et~al.(2024)Lin, Tang, Ye, Cui, Zhu, Jin, Zhang, Ning, and Yuan]{lin2024moe}
Bin Lin, Zhenyu Tang, Yang Ye, Jiaxi Cui, Bin Zhu, Peng Jin, Junwu Zhang, Munan Ning, and Li~Yuan.
\newblock Moe-llava: Mixture of experts for large vision-language models.
\newblock \emph{arXiv preprint arXiv:2401.15947}, 2024.

\bibitem[Lin(2004)]{lin2004rouge}
Chin-Yew Lin.
\newblock Rouge: A package for automatic evaluation of summaries.
\newblock In \emph{Text summarization branches out}, pp.\  74--81, 2004.

\bibitem[Lin et~al.(2023{\natexlab{b}})Lin, Zhang, Chen, Pramanick, Gao, Wang, Yan, and Shou]{lin2023univtg}
Kevin~Qinghong Lin, Pengchuan Zhang, Joya Chen, Shraman Pramanick, Difei Gao, Alex~Jinpeng Wang, Rui Yan, and Mike~Zheng Shou.
\newblock Univtg: Towards unified video-language temporal grounding.
\newblock In \emph{ICCV}, 2023{\natexlab{b}}.

\bibitem[Liu et~al.(2024)Liu, Li, Wu, and Lee]{liu2024visual}
Haotian Liu, Chunyuan Li, Qingyang Wu, and Yong~Jae Lee.
\newblock Visual instruction tuning.
\newblock In \emph{NeurIPS}, 2024.

\bibitem[Liu et~al.(2023{\natexlab{a}})Liu, Iter, Xu, Wang, Xu, and Zhu]{liu2023g}
Yang Liu, Dan Iter, Yichong Xu, Shuohang Wang, Ruochen Xu, and Chenguang Zhu.
\newblock G-eval: Nlg evaluation using gpt-4 with better human alignment.
\newblock \emph{arXiv preprint arXiv:2303.16634}, 2023{\natexlab{a}}.

\bibitem[Liu et~al.(2023{\natexlab{b}})Liu, Yang, Huang, Zhang, Huang, Wei, Deng, Sun, and Zhang]{liu2023calibrating}
Yuxuan Liu, Tianchi Yang, Shaohan Huang, Zihan Zhang, Haizhen Huang, Furu Wei, Weiwei Deng, Feng Sun, and Qi~Zhang.
\newblock Calibrating llm-based evaluator.
\newblock \emph{arXiv preprint arXiv:2309.13308}, 2023{\natexlab{b}}.

\bibitem[Lu et~al.(2024)Lu, Qiu, Ding, Zhang, Kocmi, and Tao]{lu-etal-2024-error}
Qingyu Lu, Baopu Qiu, Liang Ding, Kanjian Zhang, Tom Kocmi, and Dacheng Tao.
\newblock Error analysis prompting enables human-like translation evaluation in large language models.
\newblock In \emph{ACL}, 2024.

\bibitem[Luo et~al.(2023)Luo, Zhao, Yang, Dong, Li, Lu, Wang, Hu, Qiu, and Wei]{luo2023valley}
Ruipu Luo, Ziwang Zhao, Min Yang, Junwei Dong, Da~Li, Pengcheng Lu, Tao Wang, Linmei Hu, Minghui Qiu, and Zhongyu Wei.
\newblock Valley: Video assistant with large language model enhanced ability.
\newblock \emph{arXiv preprint arXiv:2306.07207}, 2023.

\bibitem[Maaz et~al.(2024)Maaz, Rasheed, Khan, and Khan]{Maaz2023VideoChatGPT}
Muhammad Maaz, Hanoona Rasheed, Salman Khan, and Fahad~Shahbaz Khan.
\newblock Video-chatgpt: Towards detailed video understanding via large vision and language models.
\newblock In \emph{ACL}, 2024.

\bibitem[Miech et~al.(2019)Miech, Zhukov, Alayrac, Tapaswi, Laptev, and Sivic]{miech2019howto100m}
Antoine Miech, Dimitri Zhukov, Jean-Baptiste Alayrac, Makarand Tapaswi, Ivan Laptev, and Josef Sivic.
\newblock Howto100m: Learning a text-video embedding by watching hundred million narrated video clips.
\newblock In \emph{ICCV}, 2019.

\bibitem[Min et~al.(2023)Min, Krishna, Lyu, Lewis, Yih, Koh, Iyyer, Zettlemoyer, and Hajishirzi]{min2023factscore}
Sewon Min, Kalpesh Krishna, Xinxi Lyu, Mike Lewis, Wen-tau Yih, Pang~Wei Koh, Mohit Iyyer, Luke Zettlemoyer, and Hannaneh Hajishirzi.
\newblock Factscore: Fine-grained atomic evaluation of factual precision in long form text generation.
\newblock \emph{arXiv preprint arXiv:2305.14251}, 2023.

\bibitem[Moon et~al.(2023)Moon, Hyun, Lee, and Heo]{moon2023correlation}
WonJun Moon, Sangeek Hyun, SuBeen Lee, and Jae-Pil Heo.
\newblock Correlation-guided query-dependency calibration in video representation learning for temporal grounding.
\newblock \emph{arXiv preprint arXiv:2311.08835}, 2023.

\bibitem[Munasinghe et~al.(2023)Munasinghe, Thushara, Maaz, Rasheed, Khan, Shah, and Khan]{munasinghe2023PGVideoLLaVA}
Shehan Munasinghe, Rusiru Thushara, Muhammad Maaz, Hanoona~Abdul Rasheed, Salman Khan, Mubarak Shah, and Fahad Khan.
\newblock Pg-video-llava: Pixel grounding large video-language models.
\newblock \emph{ArXiv 2311.13435}, 2023.

\bibitem[Papineni et~al.(2002)Papineni, Roukos, Ward, and Zhu]{papineni2002bleu}
Kishore Papineni, Salim Roukos, Todd Ward, and Wei-Jing Zhu.
\newblock Bleu: a method for automatic evaluation of machine translation.
\newblock In \emph{ACL}, 2002.

\bibitem[Qian et~al.(2024)Qian, Li, Wu, Ye, Fei, Chua, Zhuang, and Tang]{qianmomentor}
Long Qian, Juncheng Li, Yu~Wu, Yaobo Ye, Hao Fei, Tat-Seng Chua, Yueting Zhuang, and Siliang Tang.
\newblock Momentor: Advancing video large language model with fine-grained temporal reasoning.
\newblock In \emph{ICML}, 2024.

\bibitem[Radford et~al.(2019)Radford, Wu, Child, Luan, Amodei, Sutskever, et~al.]{radford2019language}
Alec Radford, Jeffrey Wu, Rewon Child, David Luan, Dario Amodei, Ilya Sutskever, et~al.
\newblock Language models are unsupervised multitask learners.
\newblock \emph{OpenAI blog}, 2019.

\bibitem[Radford et~al.(2021)Radford, Kim, Hallacy, Ramesh, Goh, Agarwal, Sastry, Askell, Mishkin, Clark, et~al.]{radford2021learning}
Alec Radford, Jong~Wook Kim, Chris Hallacy, Aditya Ramesh, Gabriel Goh, Sandhini Agarwal, Girish Sastry, Amanda Askell, Pamela Mishkin, Jack Clark, et~al.
\newblock Learning transferable visual models from natural language supervision.
\newblock In \emph{ICML}, 2021.

\bibitem[Raffel et~al.(2020)Raffel, Shazeer, Roberts, Lee, Narang, Matena, Zhou, Li, and Liu]{2020t5}
Colin Raffel, Noam Shazeer, Adam Roberts, Katherine Lee, Sharan Narang, Michael Matena, Yanqi Zhou, Wei Li, and Peter~J. Liu.
\newblock Exploring the limits of transfer learning with a unified text-to-text transformer.
\newblock \emph{Journal of Machine Learning Research}, 2020.

\bibitem[Regneri et~al.(2013)Regneri, Rohrbach, Wetzel, Thater, Schiele, and Pinkal]{regneri2013grounding}
Michaela Regneri, Marcus Rohrbach, Dominikus Wetzel, Stefan Thater, Bernt Schiele, and Manfred Pinkal.
\newblock Grounding action descriptions in videos.
\newblock \emph{TACL}, 2013.

\bibitem[Shen et~al.(2023)Shen, Hou, Zhou, Du, Longpre, Wei, Chung, Zoph, Fedus, Chen, et~al.]{shen2023mixture}
Sheng Shen, Le~Hou, Yanqi Zhou, Nan Du, Shayne Longpre, Jason Wei, Hyung~Won Chung, Barret Zoph, William Fedus, Xinyun Chen, et~al.
\newblock Mixture-of-experts meets instruction tuning: A winning combination for large language models.
\newblock \emph{arXiv preprint arXiv:2305.14705}, 2023.

\bibitem[Smith et~al.(2022)Smith, Patwary, Norick, LeGresley, Rajbhandari, Casper, Liu, Prabhumoye, Zerveas, Korthikanti, et~al.]{smith2022using}
Shaden Smith, Mostofa Patwary, Brandon Norick, Patrick LeGresley, Samyam Rajbhandari, Jared Casper, Zhun Liu, Shrimai Prabhumoye, George Zerveas, Vijay Korthikanti, et~al.
\newblock Using deepspeed and megatron to train megatron-turing nlg 530b, a large-scale generative language model.
\newblock \emph{arXiv preprint arXiv:2201.11990}, 2022.

\bibitem[Song et~al.(2024)Song, Chai, Wang, Zhang, Zhou, Wu, Chi, Guo, Ye, Zhang, et~al.]{song2024moviechat}
Enxin Song, Wenhao Chai, Guanhong Wang, Yucheng Zhang, Haoyang Zhou, Feiyang Wu, Haozhe Chi, Xun Guo, Tian Ye, Yanting Zhang, et~al.
\newblock Moviechat: From dense token to sparse memory for long video understanding.
\newblock In \emph{CVPR}, 2024.

\bibitem[Sun et~al.(2024)Sun, Zhang, He, Li, Cheng, Liu, Yan, Shao, Tang, Zhang, Zhao, Chen, Zheng, Zhou, Li, Zhan, Zhou, Li, Yang, Wu, Yin, Huang, Jiang, and Qiu]{Sun2024MOSS}
Tianxiang Sun, Xiaotian Zhang, Zhengfu He, Peng Li, Qinyuan Cheng, Xiangyang Liu, Hang Yan, Yunfan Shao, Qiong Tang, Shiduo Zhang, Xingjian Zhao, Ke~Chen, Yining Zheng, Zhejian Zhou, Ruixiao Li, Jun Zhan, Yunhua Zhou, Linyang Li, Xiaogui Yang, Lingling Wu, Zhangyue Yin, Xuanjing Huang, Yu-Gang Jiang, and Xipeng Qiu.
\newblock Moss: An open conversational large language model.
\newblock \emph{Machine Intelligence Research}, 2024.

\bibitem[Touvron et~al.(2023)Touvron, Lavril, Izacard, Martinet, Lachaux, Lacroix, Rozi{\`e}re, Goyal, Hambro, Azhar, et~al.]{touvron2023llama}
Hugo Touvron, Thibaut Lavril, Gautier Izacard, Xavier Martinet, Marie-Anne Lachaux, Timoth{\'e}e Lacroix, Baptiste Rozi{\`e}re, Naman Goyal, Eric Hambro, Faisal Azhar, et~al.
\newblock Llama: Open and efficient foundation language models.
\newblock \emph{arXiv preprint arXiv:2302.13971}, 2023.

\bibitem[Wang et~al.(2023{\natexlab{a}})Wang, Li, Chen, Cai, Zhu, Lin, Cao, Liu, Liu, and Sui]{wang2023large}
Peiyi Wang, Lei Li, Liang Chen, Zefan Cai, Dawei Zhu, Binghuai Lin, Yunbo Cao, Qi~Liu, Tianyu Liu, and Zhifang Sui.
\newblock Large language models are not fair evaluators.
\newblock \emph{arXiv preprint arXiv:2305.17926}, 2023{\natexlab{a}}.

\bibitem[Wang et~al.(2023{\natexlab{b}})Wang, Lv, Yu, Hong, Qi, Wang, Ji, Yang, Zhao, Song, et~al.]{wang2023cogvlm}
Weihan Wang, Qingsong Lv, Wenmeng Yu, Wenyi Hong, Ji~Qi, Yan Wang, Junhui Ji, Zhuoyi Yang, Lei Zhao, Xixuan Song, et~al.
\newblock Cogvlm: Visual expert for pretrained language models.
\newblock \emph{arXiv preprint arXiv:2311.03079}, 2023{\natexlab{b}}.

\bibitem[Zeng et~al.(2022)Zeng, Liu, Du, Wang, Lai, Ding, Yang, Xu, Zheng, Xia, et~al.]{zeng2022glm}
Aohan Zeng, Xiao Liu, Zhengxiao Du, Zihan Wang, Hanyu Lai, Ming Ding, Zhuoyi Yang, Yifan Xu, Wendi Zheng, Xiao Xia, et~al.
\newblock Glm-130b: An open bilingual pre-trained model.
\newblock \emph{arXiv preprint arXiv:2210.02414}, 2022.

\bibitem[Zhang et~al.(2023)Zhang, Li, and Bing]{zhang-etal-2023-video}
Hang Zhang, Xin Li, and Lidong Bing.
\newblock Video-{LL}a{MA}: An instruction-tuned audio-visual language model for video understanding.
\newblock In \emph{EMNLP}, 2023.

\bibitem[Zhang et~al.(2024)Zhang, Zhang, Xu, and Tao]{zhang2024vision}
Qiming Zhang, Jing Zhang, Yufei Xu, and Dacheng Tao.
\newblock Vision transformer with quadrangle attention.
\newblock \emph{IEEE TPAMI}, 2024.

\bibitem[Zhang et~al.(2022)Zhang, Roller, Goyal, Artetxe, Chen, Chen, Dewan, Diab, Li, Lin, et~al.]{zhang2022opt}
Susan Zhang, Stephen Roller, Naman Goyal, Mikel Artetxe, Moya Chen, Shuohui Chen, Christopher Dewan, Mona Diab, Xian Li, Xi~Victoria Lin, et~al.
\newblock Opt: Open pre-trained transformer language models.
\newblock \emph{arXiv preprint arXiv:2205.01068}, 2022.

\bibitem[Zhang et~al.(2019)Zhang, Kishore, Wu, Weinberger, and Artzi]{zhang2019bertscore}
Tianyi Zhang, Varsha Kishore, Felix Wu, Kilian~Q Weinberger, and Yoav Artzi.
\newblock Bertscore: Evaluating text generation with bert.
\newblock \emph{arXiv preprint arXiv:1904.09675}, 2019.

\bibitem[Zhong et~al.(2022)Zhong, Liu, Yin, Mao, Jiao, Liu, Zhu, Ji, and Han]{zhong2022towards}
Ming Zhong, Yang Liu, Da~Yin, Yuning Mao, Yizhu Jiao, Pengfei Liu, Chenguang Zhu, Heng Ji, and Jiawei Han.
\newblock Towards a unified multi-dimensional evaluator for text generation.
\newblock \emph{arXiv preprint arXiv:2210.07197}, 2022.

\end{thebibliography}
\bibliographystyle{iclr2025_conference}
\clearpage
\appendix
\section{Appendix}

\subsection{Merging Prompt} \label{sec:merging_prompt}

\begin{figure}[h]
    \centering
    \includegraphics[width=1\textwidth]{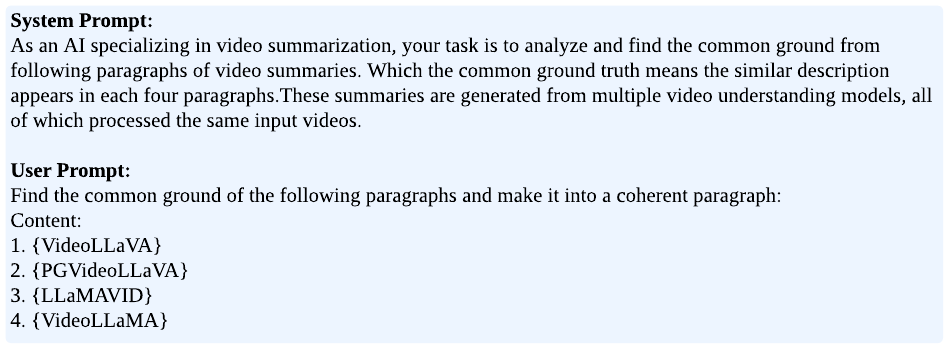}
\caption{Prompt template of ``Find common ground'' strategy in the cooperation step.}
    \label{fig:cg_prompt}
\end{figure}
\centering

\begin{figure}[h]
    \centering
    \includegraphics[width=1\textwidth]{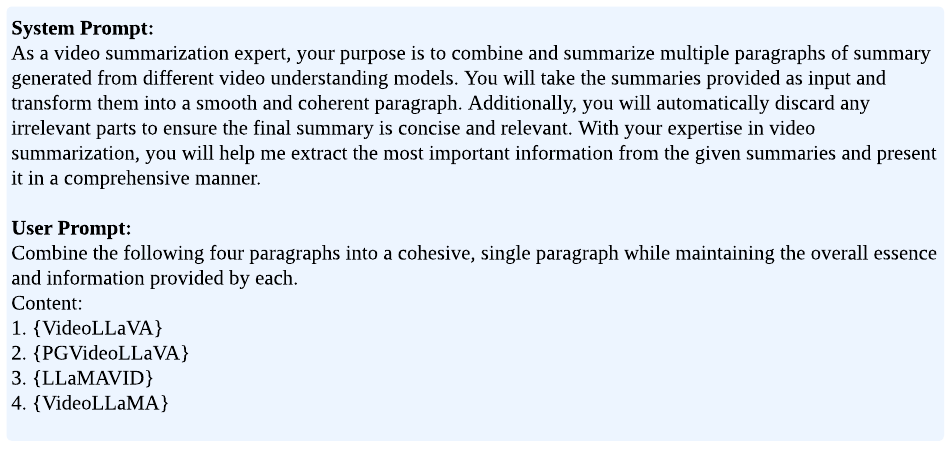}
\caption{Prompt template of ``Merge'' strategy in the cooperation step.}
    \label{fig:merge_prompt}
\end{figure}
\centering
\clearpage

\subsection{HowTo100M Qualitative Result}
\begin{figure}[H]
    \centering
    \includegraphics[width=1.0\linewidth]{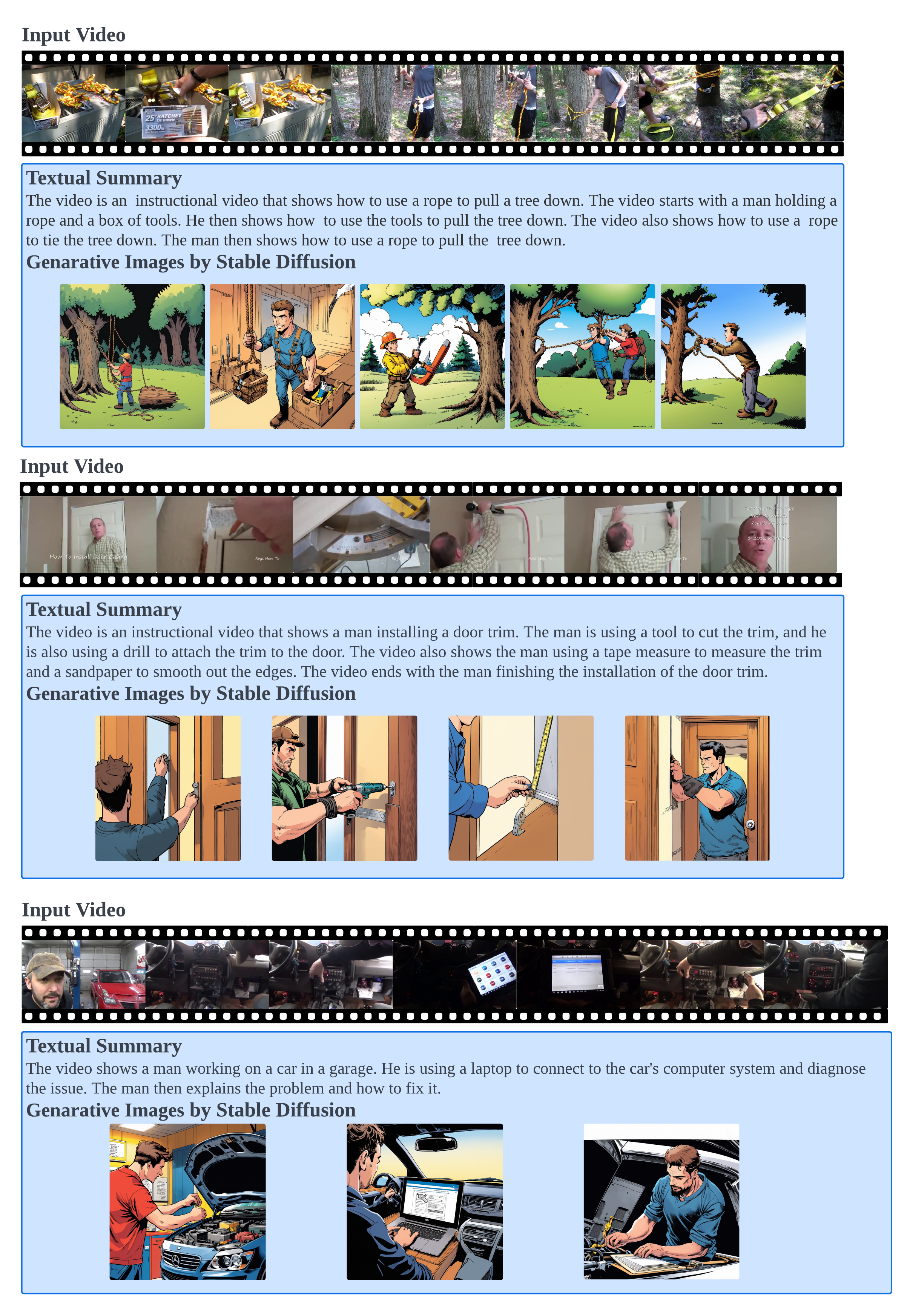}
    \caption{HowTo100M textual summary, selected keyframes, and generative results.}
    \label{fig:HowTo100M_Appendix}
\end{figure}
\clearpage

\subsection{Charades-STA Qualitative Result}
\vspace{-100pt}
\begin{figure}[h]
    \centering
    \includegraphics[width=1.1\linewidth]{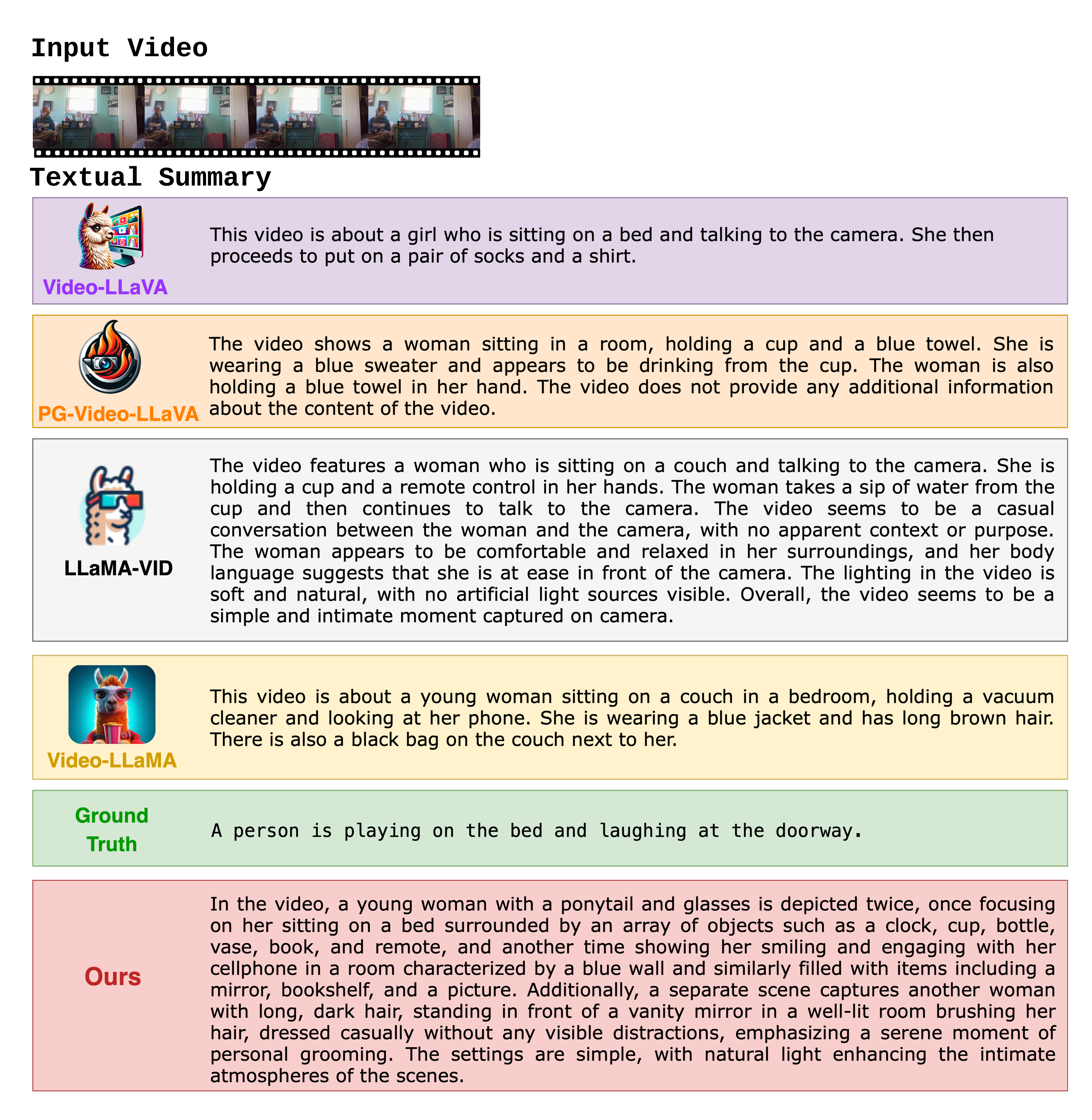}
    \caption{Textual summary qualitative comparison results on Charades-STA.}
    \label{fig:textual_charades}
\end{figure}

\subsection{TACoS Qualitative Result}
\begin{figure}[H]
    \centering
    \includegraphics[width=1.0\linewidth]{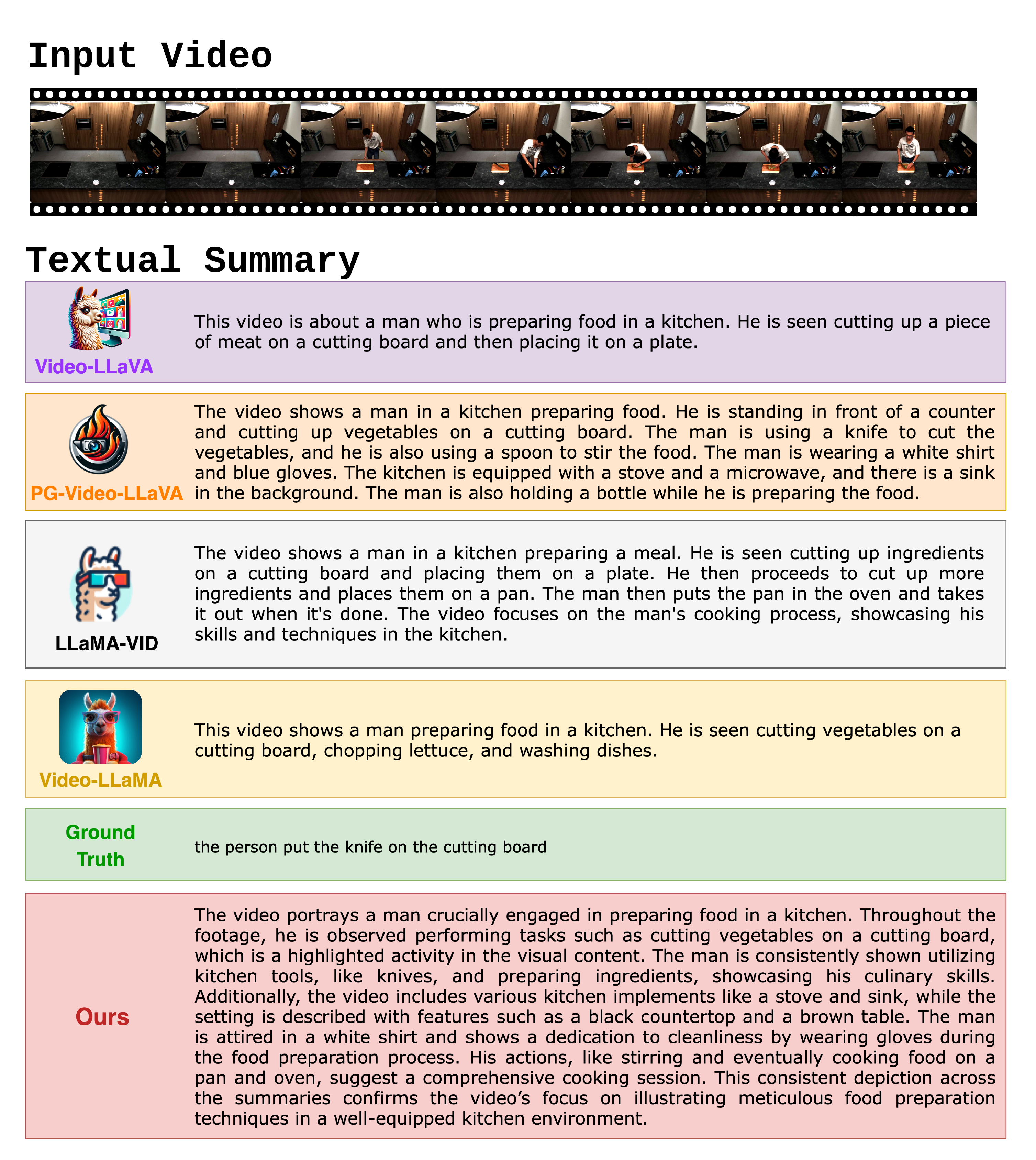}
    \caption{Textual summary qualitative comparison results on TACoS.}
    \label{fig:textual_tacos}
\end{figure}

\end{document}